\documentclass[journal]{IEEEtai}

\usepackage[colorlinks,urlcolor=blue,linkcolor=blue,citecolor=blue]{hyperref}

\usepackage{color,array}

\usepackage{amsmath,amsfonts,amssymb}
\usepackage{dsfont}
\usepackage{algorithm}

\usepackage{setspace}
\usepackage{graphicx}
\usepackage{subcaption}
\usepackage[noend]{algpseudocode}

\graphicspath{{figures/}}

\usepackage{hyphenat}

\makeatletter
\DeclareRobustCommand\onedot{\futurelet\@let@token\@onedot}
\def\@onedot{\ifx\@let@token.\else.\null\fi\xspace}

\makeatother

\begin{document}
\title{Adversary-Robust Graph-Based Learning of WSIs}
\author{Saba Heidari Gheshlaghi, Milan Aryal, Nasim Yahyasoltani, and Masoud Ganji
\thanks{ Saba Heidari Gheshlaghi and Milan Aryal contributed equally in this paper.\\ }
\thanks{ S. Heidari Gheshlaghi, M. Aryal and  N. Yahyasoltani are with the Department of Computer Science, Marquette University,
Milwaukee, WI 53202 USA (e-mail: \{saba.heidari, milan.aryal, nasim.yahyasoltani\}@marquette.edu). }
\thanks{M. Ganji is a pathologist with the Northshore Pathologists, S.C., Milwaukee, WI 53211 USA (e-mail: maganji@yahoo.com).}}

\maketitle

\begin{abstract}
Enhancing the robustness of deep learning models against adversarial attacks is crucial, especially in critical domains like healthcare where significant financial interests heighten the risk of such attacks. Whole slide images (WSIs) are high-resolution, digitized versions of tissue samples mounted on glass slides, scanned using sophisticated imaging equipment. The digital analysis of WSIs presents unique challenges due to their gigapixel size and multi-resolution storage format. In this work, we aim at improving the robustness of cancer Gleason grading classification systems against adversarial attacks, addressing challenges at both the image and graph levels. As regards the proposed algorithm, we develop a novel and innovative graph-based model which utilizes GNN to extract features from the graph representation of WSIs. A denoising module, along with a pooling layer is incorporated to manage the impact of adversarial attacks on the WSIs. The process concludes with a transformer module that classifies various grades of prostate cancer based on the processed data. To assess the effectiveness of the proposed method, we conducted a comparative analysis using two scenarios. Initially, we trained and tested the model without the denoiser using WSIs that had not been exposed to any attack. We then introduced a range of attacks at either the image or graph level and processed them through the proposed network. The performance of the model was evaluated in terms of accuracy and kappa scores. The results from this comparison showed a significant improvement in cancer diagnosis accuracy, highlighting the robustness and efficiency of the proposed method in handling adversarial challenges in the context of medical imaging.
\end{abstract}


\begin{IEEEImpStatement}
The significance of adversarial attacks and the enhancement of the robustness of deep learning (DL) models are crucial factors in establishing trust in the deployment of these technologies, particularly in sensitive domains such as healthcare. Given the significant financial incentives in the healthcare domain, the occurrence of adversarial attacks is an inevitable challenge that must be addressed to ensure a safe and reliable use of these technologies. This research explores various adversarial attacks in cancer grading scenarios using WSIs. The medical dataset, including WSIs can be attacked while stored in the cloud or during the training phase. We have addressed both cases in this work. Although there has been significant research focused on enhancing the robustness of convolutional neural network (CNN) based models, they cannot handle irregular, non-Euclidean data structures like graphs.  Deep learning models based on GNN have shown significant improvement in medical imaging, where understanding the relational information between various regions or cells within a tissue is often critical for accurate diagnosis. Through a novel method, this work shows that the integration of a GNN classifier with a denoiser and a transformer leads to improved robustness of the cancer grading, allowing it to withstand more effectively with different levels of adversarial attacks.

\end{IEEEImpStatement}

\section{Introduction}
\label{sec:introduction}
\IEEEPARstart{I}{n} recent years, DL has revolutionized various fields, especially healthcare, by enhancing diagnostics, treatment planning, and patient care. A significant impact of DL is evident in medical image analysis, including X-rays, MRIs, and WSIs \cite{ref3}. WSIs provide high-resolution digital versions of  tissue samples, offering detailed examination capabilities. Despite the benefits, DL's robustness on data patterns introduces vulnerabilities, notably adversarial attacks. These attacks involve subtle alterations to input data, designed to fool DL models into making errors. Such perturbations, often undetectable to humans, can significantly degrade a model's performance and accuracy by exploiting its dependencies on data features \cite{attack1, xu2019topology,zang2020graph,tao2020adversarial, ma2019attacking}. Furthermore, the concept of "adversarial transferability" suggests that attacks developed for one model may also compromise another, highlighting the critical challenge of ensuring robustness in DL applications \cite{papernot2016transferability}.

In the medical domain, the consequences of adversarial attacks can be particularly severe. For instance, an incorrect classification in a medical image might result in inaccurate diagnoses or treatment advice, potentially putting patients' lives or treatment plans at risk \cite{apostolidis2021survey}. Consider a situation where a deliberately engineered perturbation in a WSI of a biopsy sample causes an ML model to incorrectly identify cancerous tissue as benign or vice versa. Such an adversarial attack could lead to misguided treatment choices or unnecessary surgeries. This vulnerability raises critical concerns about the reliability of these systems in healthcare, questioning their trustworthiness when they can be easily misled. WSIs are used in pathology to identify and classify tissue abnormalities, and they are a critical component in the diagnosis of diseases like cancer \cite{advmed1,advmed2, ref3, ref4}. While the idea of adversarial attacks in the medical domain is not explored as often, there are substantial reasons for this concern. For instance, an attacker could alter test reports for personal financial gain, potentially leading to unnecessary surgeries, or these manipulations could result in incorrect diagnoses, adversely affecting patient safety and incurring unwarranted healthcare costs. The research by Apostolidis et al. \cite{apostolidis2021survey} examines the effects of adversarial attacks and defense strategies in medical imaging. Their study underscores that attacks based on gradient methods are notably prevalent and effective in the healthcare domain. Ghaffari et al. \cite{ghaffari2022adversarial} study showed the power of vision transformers against adversarial attacks in computational pathology. The study concludes that the robustness of vision transformers is based on their mechanisms of processing images, helping to have a more robust method in computational pathology against adversarial.

\begin{figure*}
    \centering
    \includegraphics[width=\textwidth]{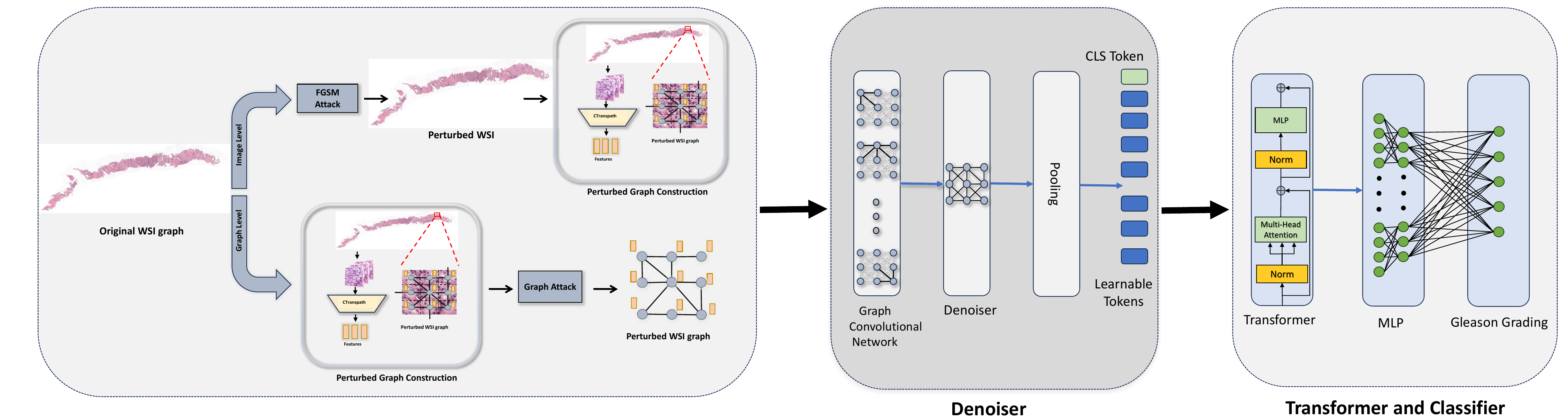}
    \caption{The overall network architecture. First, we perform data pre-processing, followed by the utilization of GNN to extract features from the graph that represent WSIs. Subsequently, these features undergo a graph denoising process and a pooling layer. Finally, the denoised graph is fed into a transformer model to classify the grades of prostate cancer. }
    \label{fig:model}
\end{figure*}

Recent studies highlight the growing use of WSIs in computational pathology and cancer detection. Most DL models, especially CNNs, face challenges in processing the vast scale of WSIs, which can reach billions of pixels and file sizes over a gigabyte. A typical solution is to divide these images into smaller patches for independent processing \cite{CNN1,CNN2, campanella2019clinical,li2021multi}. However, this approach may overlook crucial contextual information, as it fails to consider the relationships between patches. GNNs have emerged as an effective alternative to analyze WSIs and address the shortcomings of CNNs. GNNs excel at capturing spatial dependencies and complex relationships within the high-resolution structure of WSIs, crucial for accurate cancer diagnosis and classification \cite{milan2,levy2020topological,guan2022node, milan}. By integrating localized and spatial data, GNNs provide a comprehensive analysis of tissue structures, which is essential to identify detailed histopathological characteristics such as cell differentiation and tumor architecture.

As highlighted earlier, DL models are susceptible to adversarial attacks \cite{szegedy2013intriguing}, and since GNNs are a subset of DL, they inherently share this vulnerability. In recent years, there has been increasing attention to both the nature of adversarial attacks on GNNs and the development of strategies to enhance their robustness. Z\"ugner et al. \cite{advg1} pioneered an adversarial attack methodology for GNN known as Nettack. Nettack employs a surrogate model to generate adversarial samples and attack the node classification task. This technique evaluates the vulnerability of GNNs by iteratively creating adversarial samples and adjusting these samples based on the changes in confidence values resulting from generated adversarial samples. Although this method is known as an early and influential method of adversarial attack against GNNs, its practicality is hindered for large-scale graphs due to time complexity.
Entezari et al. \cite{entezari2020all} employed the first-gradient optimization method in order to address the bi-level optimization problem and generate adversarial samples. Furthermore, this paper used a low-rank approximation strategy to reduce the impact of the attack and enhance the network's robustness. Another gradient-based approach, named projected gradient descent (PGD), was conducted by Chen et al. \cite{chen2018fast}, which involves computing the gradients for the edges and identifying those with the highest absolute values between the pair of nodes to inform their attack strategy. Sun et al. \cite{sun2020adversarial}  applied a surrogate model to address the bi-level optimization challenge and employed meta-gradient methods to manage the addition or removal of edges. Their findings also indicated a slight tendency of the algorithm toward connecting nodes of lower degrees. Dai et al. \cite{dai2018adversarial} used a reinforcement learning approach to generate adversarial samples to disrupt node classification tasks. Zang et al. \cite{zang2020graph} concentrated on identifying "bad actor" nodes and showed that by flipping the edges between such nodes and a chosen target node, the overall network's performance can be negatively impacted.

Ensuring the robustness of DL models against adversarial attacks is critical, especially in the medical field, where accuracy in diagnosis and patient safety is of paramount importance. Robust DL models help reduce diagnostic errors and the risk of false results, thus maintaining high standards of patient care. With the rise of medical image storage on cloud platforms, accessibility for healthcare professionals has improved, but this has also introduced significant security challenges. Medical images in the cloud are vulnerable to cyber-attacks, potentially leading to unauthorized modifications that compromise data integrity and system performance.

Furthermore, DL models analyzing these images are susceptible to adversarial attacks, particularly when images are converted to graphs for analysis. Such vulnerabilities can degrade the models' performance and reliability. {This paper addresses the security threats to medical images in cloud storage into two different ways: image-level attacks and graph-level attacks. Image-level attacks refer to direct alteration or perturbation of the images themselves, compromising their authenticity and reliability. On the other hand, graph-level attacks involve the manipulation of the images after they have been converted into graph representations for analysis, targeting the structural and relational integrity of the data. This differentiation helps in understanding the nature of the threats and devising specific countermeasures for each type of attack, aiming to enhance the security and robustness of medical images storage and processing systems.} 

Adversarial training is a key method to improve model robustness, where adversarial examples are intentionally used during training to help models learn robust and discerning features \cite{tramer2017ensemble, loveland2021reliable}. Studies have also investigated the use of advanced normalization techniques and architectural modifications to mitigate the effects of adversarial attacks. Additionally, inverse imaging problems (IIPs) have shown promise in reducing input noise and perturbations, though their effectiveness often depends on having paired noise and clean data, a limitation addressed by the introduction of untrained neural network priors (UNNPs) for noise reduction without pre-existing data \cite{qayyum2022untrained, bora2017compressed}. Luo et al.  \cite{luo2021learning} developed a model to improve GNN robustness by learning to eliminate irrelevant edges and reduce graph complexity. Graph attention networks (GATs) further enhance the robustness of GNN by focusing on crucial nodes or edges, minimizing noise impact \cite{GAT}.
\section{Method}
\label{sec:Method}
\begin{figure}[b]
\includegraphics[width=\linewidth]{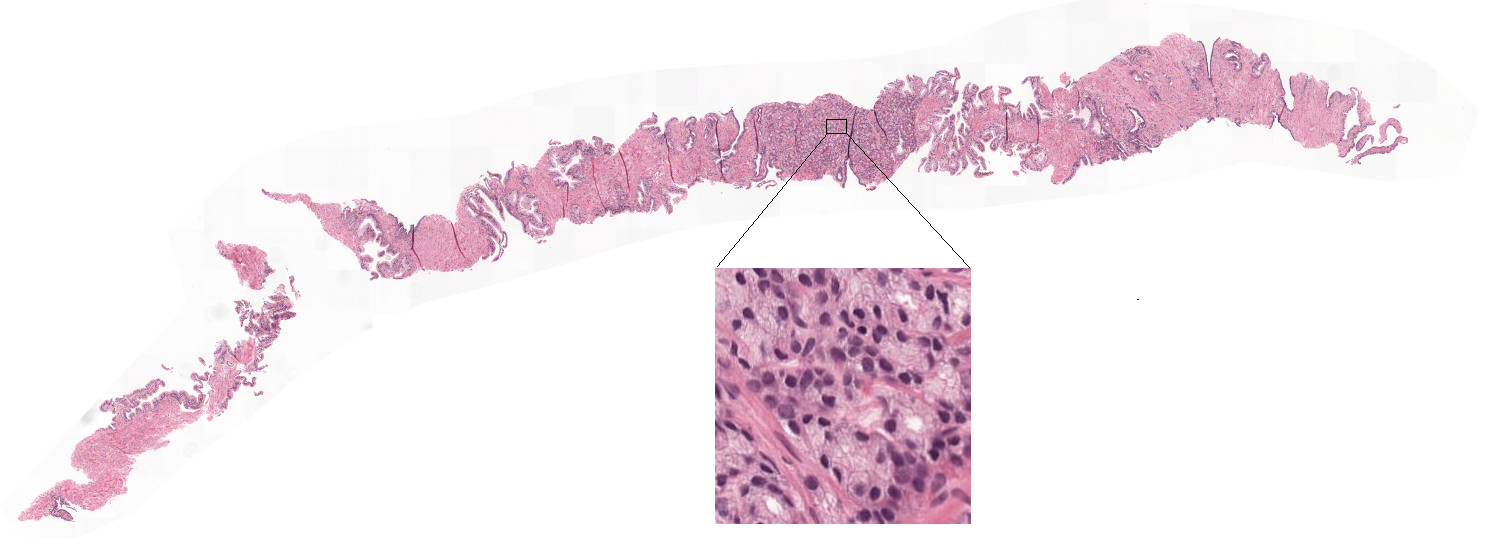}
\centering
\caption{WSI without any perturbation.}
\label{fig:original}
\end{figure}
 As regards cancer detection using WSIs, research has mainly focused on CNNs for patch-based image analysis. Foote et al. \cite{foote2021now} identified vulnerabilities of the DL model to universal perturbations that could alter tumor image classifications. Ghaffari et al. \cite{ghaffari2022adversarial} showed CNN susceptibility to adversarial attacks, countering this with adversarial training and dual batch normalization (DBN). They found that vision transformers (ViTs) inherently resist adversarial attacks without needing specific pretraining or architectural changes. Bhojanapalli et al. \cite{bhojanapalli2021understanding} highlighted ViT robustness, which persists even after removing a layer during training with sufficient data. Zhang et al. \cite{zhang2022benchmarking} evaluated CNN robustness against common WSI corruptions, using classification and ranking metrics to assess their vulnerability.

In this paper, we present a novel GNN-based architecture for digital pathology that stands out for its unique approach to enhancing robustness against adversarial attacks in WSIs. This work is distinguished by its two main factors: first, it is the pioneering effort specifically designed to protect models from adversarial threats targeting both image and graph representations of WSIs; second, it moves beyond traditional patch-level analysis, addressing the need for context-aware evaluation around tumor areas in WSIs. Prior research has shown graph-based methods to be more effective than patch-based approaches \cite{graph1,graph2,graph3}. The proposed model showcases superior robustness to various levels of adversarial perturbations in WSIs. By generating and testing against adversarial attacks, we benchmark the robustness of the proposed architecture against leading GNN models. The findings confirm the method's outstanding capability to mitigate challenges in clinical applications. The core contributions of this study include:
\begin{enumerate}
\item Evaluating the susceptibility of cutting-edge GNN models to adversarial threats in WSIs.
\item Introducing a novel graph-based architecture designed to enhance robustness against adversarial threats in WSIs, particularly focusing on cloud-stored and graph-converted WSIs.
\end{enumerate}

In this section, we provide a comprehensive explanation of the methodologies used in this study. Figure \ref{fig:model} shows the overall architecture of the proposed algorithm. We begin by elucidating the concepts of GNN and graph-based learning. Given that WSIs are stored on cloud platforms, they are vulnerable to adversarial attacks at various stages. These attacks may target the WSIs directly or occur during their conversion to graph representations for training purposes. In light of these vulnerabilities, this work encompasses both graph- and image-level attacks, providing a detailed examination of each type. Finally, we provide a detailed overview of the transformer and denoiser components involved in the proposed robust network. 
\subsection{Graph-Based Learning}
Graph-Based Learning is an innovative approach in the DL domain, focusing on graph-based data. A graph consists of nodes ($V$) and edges ($E \subseteq v \times v$), where the nodes represent entities, and the edges denote the relationships between these entities. In addition, it can be represented by the adjacency matrix$A \in \mathbb{R}^{N \times N}$ and the features$\boldsymbol{x}\in \mathbb{R}^{N}$ and defined as $G(V,A,\boldsymbol{x})$. GNN is effective in handling complex relational data, which is not easily manageable by conventional DL models. In this research, initially, each WSI is divided into non-overlapping patches, from which features are extracted for each individual patch. Pretrained CTranspath \cite{wang2022} has been used to extract the features. In this graph-based representation, each patch is treated as a node, and the edges represent the spatial relationships or contextual dependencies between these patches. that are established using the k-nearest neighbors (k-NN) algorithm. This process results in the construction of an adjacency matrix that defines the connections between nodes, effectively representing the WSI as a graph. With WSIs converted into graphs, GNNs can be applied to classification tasks. In GNNs, the learning process involves message passing between adjacent nodes \cite{kipf2017semisupervised}. The hidden layer \(l+1\) is defined as:
\begin{equation}
H^{(l+1)}=\sigma\left(\tilde{D}^{-\frac{1}{2}} \tilde{A} \tilde{D}^{-\frac{1}{2}} H^{(l)} W^{(l)}\right)
\end{equation}
Where for the layer $ls$, $H^{(l)} \in \mathcal{R}^{|V|\times d}$ represents the input, where each node possesses $d$ features. The trainable weight matrix at this layer is denoted as $W^{(l)} \in \mathcal{R}^{|V|\times d} $. The non-linear activation function is represented by $\sigma$. The normalized symmetric adjacency matrix is given by $\hat{A}:=\tilde{D}^{-\frac{1}{2}}\tilde{A}\tilde{D}^{-\frac{1}{2}}$, where $\tilde{A} := A + I$ signifies the adjacency matrix enhanced with self-connections. $\tilde{D}$ is the diagonal matrix corresponding to $\tilde{A}$ with $\tilde{D}_{ii} :=  \sum_j \tilde{A}_{ij}$ representing the diagonal elements. Hence, GNNs with a single hidden layer are defined as
\begin{equation}
Z=f_{\theta}(A, X)=\operatorname{softmax}\left(\hat{A} \sigma\left(\hat{A} X W^{(1)}\right) W^{(2)}\right)
\end{equation}
Where \(\theta=\left\{W^{(1)}, W^{(2)}\right\}\) is the set of parameters that can be learned by minimizing cross-entropy on the output of the labeled samples \(\mathcal{V}_{L}\):
\begin{equation}
L(\theta ; A, X)=-\sum_{v \in \mathcal{V}_{L}} \ln Z_{v, c_{v}}, \quad \quad \quad Z=f_{\theta}(A, X)
\end{equation}
where \(c_{v}\) represents the label assigned to node \(v\) of the training data set. Following completion of the training process \(Z\) means the probabilities associated with class for each instance within the graph.
An adversarial sample is an input meticulously engineered to deceive a DL model, resulting in the misclassification of the input. To the human eye, these perturbed inputs may appear identical to their original forms, yet they cause the network to misinterpret the contents of the input. Adversarial attacks can occur in both images and graphs. In this section, we will discuss the methods through which these attacks can occur in either image-based or graph-based data. { Figures \ref{fig:original} shows a sample WSI with no perturbation. The goal is to evaluate the effects of adversarial attacks on these images.
\begin{figure*}
\centering
\begin{subfigure}[b]{.8\textwidth}
\includegraphics[width=\textwidth]{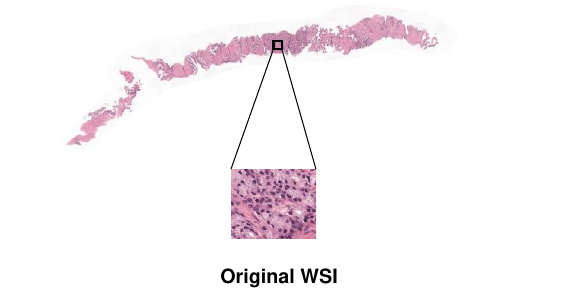}
\label{fig:mouse}
\end{subfigure}

\begin{subfigure}[b]{.49\linewidth}
\includegraphics[width=\textwidth]{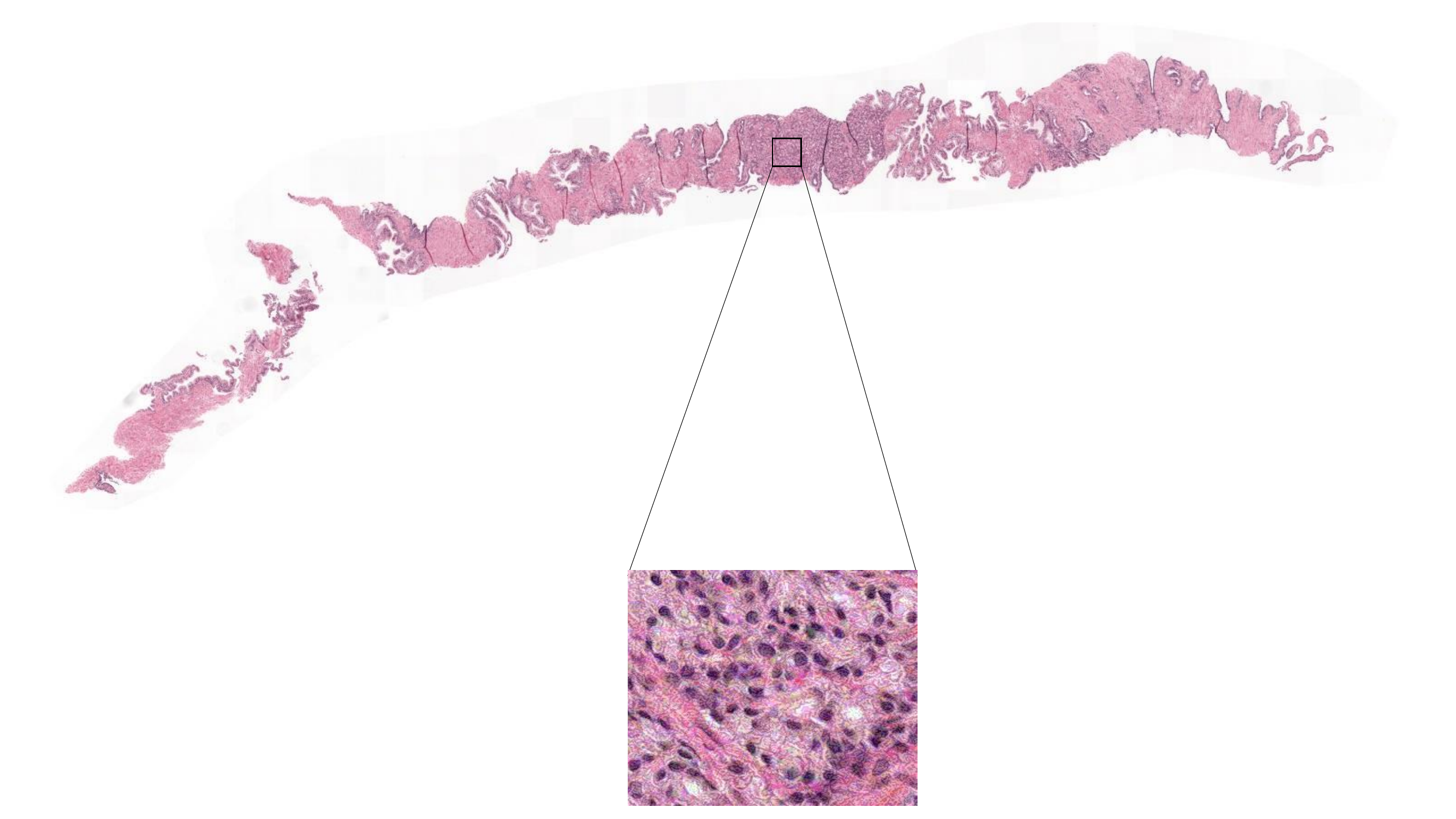}
\caption{$\epsilon = 0.1$}\label{fig:gull}
\end{subfigure}
\begin{subfigure}[b]{.49\textwidth}
\includegraphics[width=\textwidth]{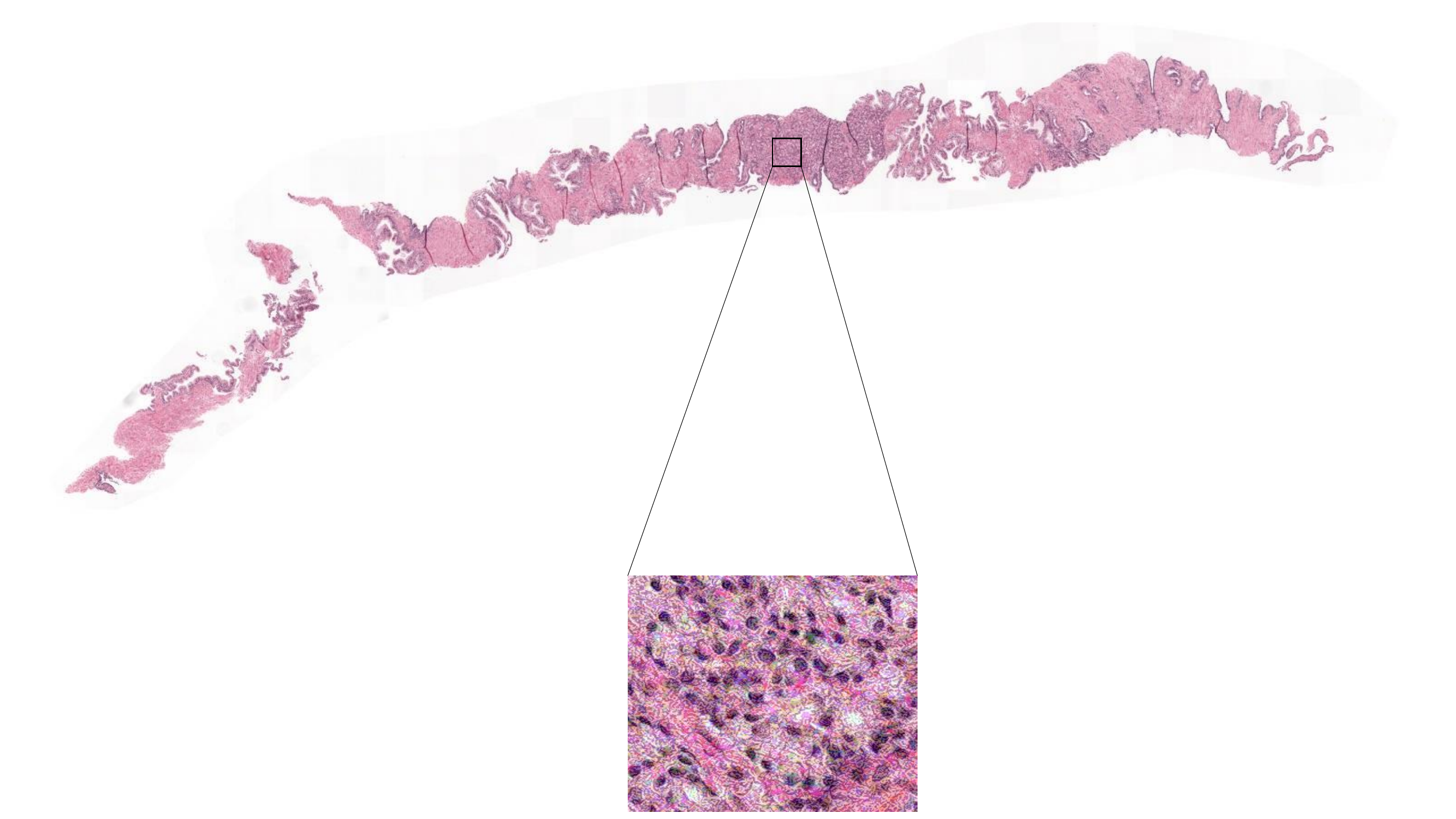}
\caption{$ \epsilon = 0.15$}\label{fig:tiger}
\end{subfigure}
\caption{WSI with FGSM adversarial attack.}
\label{fig:attack}
\end{figure*}

The Fast Gradient Sign Method (FGSM) \cite{attack1} is a powerful and popular adversarial attack technique designed to fool ML models, especially DL networks.  Noted for its simplicity and effectiveness, the FGSM attack has become a fundamental concept in the study of adversarial attacks, providing key insights into the vulnerabilities of ML systems. The FGSM attack exploits the vulnerability of ML models to small, carefully crafted perturbations in the input data. Its key idea is to compute the gradient of the model's loss with respect to the input data and then add a small perturbation to the input in the direction that maximizes the loss. This perturbation is determined by the sign of the gradient.  In other words, the attack uses the gradient of the loss with respect to the input data and then adjusts the input data to maximize the loss.
The process can be broken down into the following steps:
\begin{itemize}
\item Compute the gradient of the loss with respect to the clean and unperturbed input image.
\item Apply a small perturbation to the input in the direction of the gradient. This perturbation is usually done by adding the sign of the gradient to the original input.
\item The resulting is an adversarial sample.
\end{itemize}
Mathematically, for a given input data point $x$ and a ML model with a loss function $J$ and parameters $\theta$, the FGSM attack generates an adversarial example $x_{adv}$ with perturbation budget of $\epsilon$ as follows:

\begin{equation}
    x_{adv}= x + \epsilon . sign (\nabla_x J(\theta,x,y)) 
\end{equation}

Here, $\epsilon$ represents a small value selected to fulfill the criteria of being unnoticeable, such that ($x\cong x_{adv}$).\\
Figure \ref{fig:attack} illustrates the original image and the effects of an image-level attack, respectively. More specifically,  the impact of the FGSM attack at various attack perturbation rates is demonstrated.}



\begin{figure}[t]
\includegraphics[width=\linewidth]{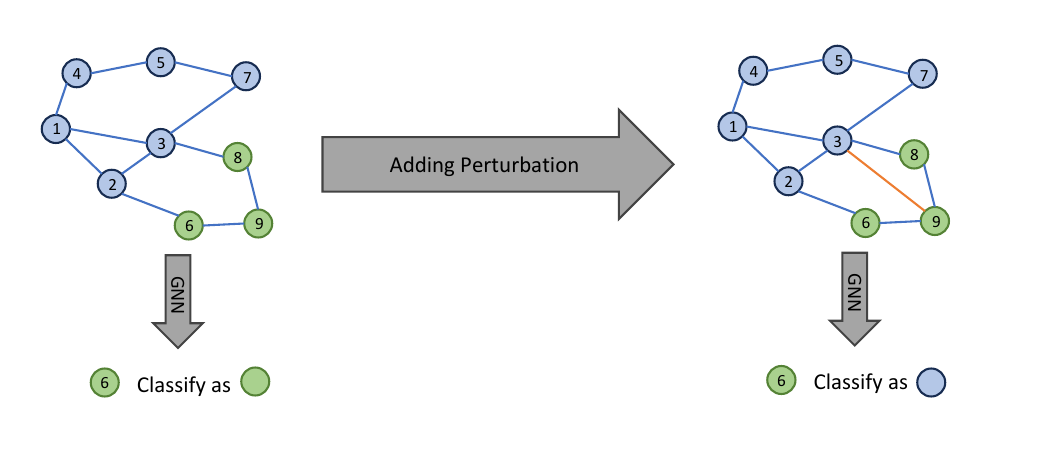}
\centering
\caption{Attacker's effect on network classification.}
\label{fig:adv_graph}
\end{figure}

\subsubsection{Graph-level attack}
Adversarial attacks are categorized into various types, each distinguished by specific characteristics such as the level of access the attacker has, the attacker's goal, and the attacker's knowledge about the target system. This categorization helps us to understand the nature of adversarial attacks in the graph domain and their potential impact on different models and knowledge.

Adversarial attacks in the graph domain present a higher level of complexity compared to other domains, such as images. This increased challenge stems from the broader array of attack types and options available when targeting graph-based models. The methods of perturbation in graphs are diverse and include various tactics. These include:

\begin{itemize}

\item \textbf{Modifying features:} The attacker is capable of changing the node's features without affecting the overall structure of the graph.

\item \textbf{Adding or deleting edges:} The attacker is capable of adding or removing edges between nodes. Figure \ref{fig:adv_graph} is an example of how the attacker can affect the network performance by adding an edge.

\item \textbf{Injecting or removing nodes:} The attacker can insert a fake node or add the existing node to the graph in order to affect the network performance.

\end{itemize}

Attackers can possess varying levels of access and capabilities, which often depend on the stage at which the attack occurs. Their access can be categorized into :

\begin{itemize}
\item \textbf{Evasion attack:} Evasion attack happens post-training or during the testing phase of a model. In this type of attack, the model's parameters are already set and fixed, meaning that the attacker is unable to change or modify them. Instead, the attacker focuses on manipulating the input data in such a way that the model is fooled and makes incorrect predictions or classifications. 

\item \textbf{Poisoning attack:} Poisoning attack happens before the model is trained. In this type of attack, the attacker poisons the model's training data in such a way that the model is fooled and makes incorrect predictions or classifications \cite{tang2020transferring, liu2019unified, zhou2019data, zhang2019towards}.  

\end{itemize}

Attackers may have a variety of objectives and goals during an attack, each with its own specific intent and impact. Attackers goals can be:

\begin{itemize}
\item \textbf{Targeted attack:} Attacker's primary objective is to manipulate the classification of a subset of test nodes. In this type of attack, the attacker targets precisely a specific node (or nodes) within a network or dataset, aiming to alter their classification outcomes.

\item \textbf{Untargeted attack:} Attacker’s primary objective is to manipulate and affect the overall performance of the network. In this type of attack, the attackers do not have any targeted node and aim to alter the overall classification outcome.
\end{itemize}

The level of knowledge that attackers have can be different. Some attackers might have extensive access to detailed information, while others may only have limited insights. This variability in access and knowledge directly impacts the strategies and methods they can employ in their attacks. Attacker's knowledge can be categorized into:

\begin{itemize}

\item \textbf{White-box attack:} The attacker has access to the full knowledge and various key components, including the model, network parameters, gradient information, training inputs, and labels, among other elements. This level of access grants them a significantly advanced knowledge base. Such extensive insight enables the attacker to execute sophisticated and potentially more effective attacks, utilizing their deep understanding of the system's inner workings. However, it is important to note that this kind of attack, with its requirement for extensive access and knowledge, is often not feasible and realistic in many real-world scenarios.
\item \textbf{Grey-box attack:} The attacker has access to limited information about the victim's model, making this attack potentially more dangerous than a white-box attack. For instance, the attacker can access the model's parameters but not its training data.

\item \textbf{Black-box attack:} The attacker is unable to access the model, its parameters, or the training labels. Instead, their capabilities are limited to accessing only the adjacency matrix or features, and conducting black-box queries to obtain output scores or labels  \cite{wu2019adversarial}.
\end{itemize}

In the context of graph-level attacks targeting the discrete structure of graphs, the objective is to develop a function $f_{\theta}$, where $\theta$ is typically determined by minimizing an attack loss function,  $\mathcal{L}_{\text {atk }}$, applied to the labeled training nodes of the graph $G$. This process can be succinctly described as a bilevel optimization challenge where $\hat{G}$ is the perturbed graph.

\begin{equation}
    \begin{aligned}
        \min _{\hat{G} \in \Phi(G)} \mathcal{L}_{\text {atk }}\left(f_{\theta^{*}}(\hat{G})\right) \quad \text { s.t. } \quad \\
        \theta^{*}=\underset{\theta}{\arg \min } \quad \mathcal{L}_{\text {train }}\left(f_{\theta}(\hat{G})\right) .
    \end{aligned}
\end{equation} 
Where a set of admissible perturbations on the data is $\Phi(G)$, where $G$ is the graph at hand.

To address the bilevel optimization challenge, we adopt a meta-learning approach, initially presented in Zügner et al.'s work \cite{zügner2019adversarial}. Meta-learning, often referred to as "learning to learn," is a method designed to enhance the efficiency of ML models, particularly in terms of data and time resources \cite{andrychowicz2016learning}. In the context of an adversarial attack on GNNs, the matrix representing the graph's structure is considered a hyperparameter and the meta-gradient is computed with respect to the graph structure matrix. The primary objective is to determine the gradient of the attacker's loss following the training phase in relation to this hyperparameter and this computation involves backpropagation through the learning phase of the model. The attacker's loss after training $\mathcal{L}_{\text {atk }}\left(f_{\theta^{*}}(G)\right)$, is used to compute the meta-gradient, $\nabla_{G}^{\text {meta }}$, which shows how this loss changes with small perturbations in the graph data. This calculation of the gradient is based on the following, where $opt(.)$ represents a differentiable optimization process, such as gradient descent or its stochastic variations.

\begin{equation}
    \begin{aligned}
        \nabla_{G}^{\text {meta }}=\nabla_{A} \mathcal{L}_{\text {atk }}\left(f_{\theta^{*}}(A)\right) \quad \text { s.t. } 
        \\
        \theta^{*}=\operatorname{opt}_{\theta}\left(\mathcal{L}_{\text {train }}\left(f_{\theta}(A)\right)\right),
    \end{aligned}
\end{equation}

In this method, we focus on executing adversarial attacks that disrupt the discrete graph structure. The objective is to modify the original graph structure \(G^{(0)}=\left(A^{(0)}, X^{(0)}\right)\) into a modified version $\hat{G}=(A^{\prime}, X^{(0)})$ in order to impair the classification performance. This approach is a non-targeted poisoning attack, which targets the graph structure, while node attributes are treated as constant. In this approach, a greedy strategy is used for choosing graph perturbations, such as edge insertions or deletions, based on their anticipated influence on achieving the attacker's objectives.

However, Computing meta-gradients is computationally expensive. To address this, a first-order approximation proposed by Finn et al. has been used \cite{finn2017model}. This approximation is based on taking the gradient of the attack loss with respect to the data after training the model for a certain number of steps. It does not consider the training dynamics in its calculation, making it a less expensive computational alternative. Using this approximation, we have:

\begin{equation}
    \begin{aligned}
        \nabla_{A}^{\text {meta }}=\nabla_{A} \mathcal{L}_{\text {atk }}\left(f_{\theta_{T}}(A)\right)
        \\
        \approx \nabla_{A} \mathcal{L}_{\text {atk }}\left(f_{\tilde{\theta}_{T}}(A)\right)
        \\
        =\nabla_{f} \mathcal{L}_{\text {atk }}\left(f_{\tilde{\theta}_{T}}(A)\right) \cdot \nabla_{A} f_{\tilde{\theta}_{T}}(A)
    \end{aligned}
\end{equation}

Where $\tilde{\theta}_{t}$ represents the parameters at time $t$, which are independent of the data $A$, meaning that 
$\nabla_{A} \tilde{\theta}_{t}=0$. Consequently, the gradient does not extend through $\tilde{\theta}_{t}$. This is equivalent to computing the gradient of the attack loss, $\mathcal{L}_{\text {atk }}$, with respect to the data after the model has been trained for $T$ steps. Additionally, a heuristic approach to meta-gradients is explored, which updates the initial weights based on the direction that typically results in the largest increase in training loss. This heuristic is more efficient to compute and has a smaller memory footprint compared to the exact meta-gradient. The attack algorithm uses this heuristic by combining both the training loss and a self-learning estimated loss.\\

\subsection{Transformer}

Transformers, initially created for tasks in natural language processing, have now been adapted for use in GNNs. This evolution marks a significant advancement in managing graph-structured data and pinpointing long-range dependencies within it. Integration of transformers with GNN opens up new avenues for processing and interpreting complex data structures \cite{yun2019graph}. This concept draws inspiration from the work presented in the literature, specifically in the studies by Vaswani et al. \cite{vaswani2023attention} and Zheng et al.~\cite{zheng2022graphtransformer}, which delve into the mechanics of attention mechanisms and their application to graph-transformer models. These studies exemplify the potential of combining transformer models with GNNs, demonstrating how they can complement each other in enhancing data analysis, especially in scenarios involving intricate and interconnected data. This synergy has the potential to lead to more robust and effective solutions in the field of data science and analytics.

In the transformer model, the input is denoted as $H^{pool} \in \mathcal{R}^{N \times d}$, where  $N$ is the total count of nodes within the network, and $d$ stands for the dimensionality of the features that have been processed through the graph's pooling layer. This structure is designed in such a way that each node, accompanied by its corresponding set of features, is considered as an individual input token for the transformer. This approach allows the transformer to interpret and process each node and its associated features distinctly, contributing to the model's ability to understand and analyze complex graph structures effectively. Furthermore, a class token (CLS), designed to aggregate information from the entire graph for making graph-level predictions, is also integrated into the input of the transformer, accompanying the individual node-feature tokens. The transformer itself is composed of several layers, with each layer functioning as follows:
\begin{equation}
    t_i^{'} = MHA(LN(t_{i-1}))+ t_{i-1}
\end{equation}
\begin{equation}
    t_i = MLP(LN(t^{'}_i))+t_i^{'}
\end{equation} \\
Where $MHA$ is the multi-headed self-attention mechanism, $LN$ denotes layer normalization and notation $t_i$ indicates the $i^{th}$ layer of the transformer. The initial layer, $t_0$, is given by  $t_0 = [CLS;H^{pool}]$], where $[CLS;H^{pool}]$ denotes the concatenation of the class token with the pooled node features. This layered structure of the transformer, with its attention mechanisms and normalization processes, plays a crucial role in effectively capturing and interpreting the complex patterns in the input data, ultimately leading to a robust and accurate classification process.\\

After the transformer processes the data, the output is then fed into a multi-layer perceptron (MLP). This stage is essential for achieving the final classification. The MLP, with its multiple layers, takes the output from the transformer and uses it to compute the final predicted output, represented as $\hat{y}$, for each instance of WSIs.

\begin{equation} 
    \hat{y} = MLP(Transformer([CLS;H^{pool}])
\end{equation}
The transformer consists of multiple layers and each layer in the transformer is given by,
\begin{equation}
    t_i^{'} = MHA(LN(t_{i-1}))+ t_{i-1}
\end{equation}
\begin{equation}
    t_i = MLP(LN(t^{'}_i))+t_i^{'}
\end{equation} \\
Where $LN$ refers to the layer normalization, and $t_i$ is the $i^{th}$ layer of the transformer. The transformer architecture employs a Multi-Head Attention ($MHA$) mechanism, which is a pivotal component enhancing its capability to process and understand data. The initial layer for the transformer is $t_0 = [CLS;H^{pool}]$ and $MHA$ calculated as follows:
\begin{equation}
    MHA = Concat[A_1,A_2,\cdots,A_h]W
\end{equation}
In this context, $W$ represents a matrix of trainable parameters, while $A_i$ denotes the  $i^{th}$ instance of self-attention. The self-attention mechanism operates on the principles of key(K), query(Q), value(V) mechanism which is given as follows\\
\begin{equation}
    A_i(Q,K,V) = softmax \left(\frac{QK^T}{\sqrt{d_k}}\right)V
\end{equation}\\
where $d_k = d/h$. With $W_Q,W_V,W_K \in \mathcal{R}^{d \times d_k}$ as learnable weight parameters, $Q,K,V$ used in attention follow\\
\begin{equation}
    \begin{aligned}
      Q = H^{pool}W_Q  \\
      K = H^{pool}W_K \\
      V = {H^{pool}{W_V}}.
    \end{aligned}
\end{equation}
\subsection{Denoising}

A denoising module is used for eliminating unwanted disturbances or inconsistencies from data. Specifically in the realm of adversarial attack S, this process focuses on recognizing and reducing the impact of adversarial samples. In graph domain, these attacks can manipulate the graph's structure by adding or removing edges, or they can alter the nodes or edges features. The objective of denoising in GNNs is to preserve the network's functionality and accuracy, even when faced with these adversarial challenges. By successfully cleaning the data of these perturbations, GNNs are better equipped to maintain robust and reliable performance, ensuring that predictive capabilities are not compromised by such noise and attacks.

In this study, the aim is to distinguish and remove the noise component, represented as {$\boldsymbol{n}\in \mathbb{R}^{N}$}, from the input signal {$\boldsymbol{x}\in \mathbb{R}^{N}$} and accurately estimate the denoised signal, denoted as  {$\boldsymbol{x_d}\in \mathbb{R}^{N}$}.

\begin{equation}
\boldsymbol{x} = \boldsymbol{x_d} + \boldsymbol{n}
\end{equation}

The task involves determining the optimal GNN weights denoted as $\boldsymbol{\boldsymbol{\theta}}$, by aiming to minimize the specified loss function.

\begin{equation}
\label{eq9}
l(\boldsymbol{x},\boldsymbol{\boldsymbol{\theta}}) = \frac{1}{2} || \boldsymbol{x} - f_{\boldsymbol{\theta}}(\boldsymbol{z}|G)||^2_2
\end{equation}

where  $f_{\boldsymbol{\theta}}(\boldsymbol{z}|G)$ is GNN which is a parametric non-linear function, $G$ is the graph, $\boldsymbol{\theta}$ indicates the network parameter, and $\boldsymbol{z}$ is the initial random value sourced from a Gaussian distribution with zero mean. Utilizing the principles of graph signal processing and the untrained neural network process (UNNP) framework \cite{rey2022untrained, qayyum2022untrained}, this module is designed to denoise graph inputs through the use of an untrained network. This approach optimizes the network parameters exclusively based on the individual signal observation requiring denoising, rather than using a diverse training set of multiple observed graph signals. The proposed method's aim is to precisely separate and recover the denoised signal from the noise.

In this method, we utilize the inherent potential of overparameterized networks, which are characterized by their extensive number of parameters. Furthermore, this approach incorporates the technique of early stopping. This strategy is particularly effective as it enables the network to rapidly adjust to and learn from the actual signal, while simultaneously being less susceptible to the influences of noise. This is in contrast to more traditional methods, which often take longer to differentiate between signal and noise. By applying this methodology, we aim to extract a denoised version of the graph, denoted as ($\boldsymbol{{x}'_d}$) from the given input.

The stochastic gradient descent (SGD) method enhanced with early stopping is utilized to address the minimization problem presented in \eqref{eq9}. We initiate this process by randomly setting the values of parameters  $\boldsymbol{\theta}$ and $\boldsymbol{z}$, selecting from an i.i.d zero-mean Gaussian distribution. The weights, refined over several denoising iterations on $x$ are denoted as ${\boldsymbol{\theta}}'$. This helps to have $\boldsymbol{{x}'_d}$ as the output, corresponding to these refined weights. This approach stands in contrast to the conventional method, where $\boldsymbol{\theta}$ is initially calibrated by fitting it to a training dataset.

\begin{equation}
\boldsymbol x'_d = f_{{\boldsymbol\theta'}_{(\boldsymbol x)}}(\boldsymbol Z|G)
\end{equation}

This implies that for every combination of a noisy signal, denoted as $x$ and its corresponding denoised counterpart, labeled $\boldsymbol{{x}'_d}$, a unique set of weights is inherently associated with each pair. These specific weights play a crucial role in the denoising process, as they facilitate the extraction of the most effective denoised output. This optimized output is derived directly from the input graph, ensuring that the denoising process is tailored specifically to the characteristics of each individual signal pair, thus enhancing the accuracy and efficiency of the denoising operation.\\

\begin{table}[h]
\caption{Number of samples of GG in prostate cancer dataset}
\begin{center}
\begin{tabular}{|c|c|}
\hline
\textbf{Description }& \textbf{Number of samples} \\
\hline
GG1 & 2666  \\
\hline
GG2 & 1343  \\
\hline
GG3 & 1250\\
\hline
GG4 & 1242  \\
\hline
GG5 & 1224\\
\hline
\end{tabular}
\label{tab:pca}
\end{center}
\end{table}

\section{Dataset Description}

The PANDA (Prostate cANcer graDe Assessment) dataset is an important resource in the medical field, specifically designed for the accurate and efficient grading of prostate cancer. It consists of prostate cancer biopsy samples classified into five distinct grades based on the Gleason scoring system. The significance of this dataset is underscored by the widespread incidence of prostate cancer, which is globally recognized as the second most prevalent cancer in men. Alarmingly, this disease is estimated to cause around 350,000 deaths annually \cite{bulten2022artificial}. Encompassing roughly 10,000 WSIs, the PANDA dataset provides an extensive foundation for medical research and analysis. Comprehensive details about the contents and characteristics of each sample in this dataset are methodically presented in Table~\ref{tab:pca}.

To further enrich the dataset’s value, about 25\% of these WSIs have been subjected to a random selection process. Post-selection, these slides undergo the introduction of various perturbations. This deliberate modification is a critical aspect of the dataset, as it simulates adversarial attacks that medical professionals may encounter in diagnosing and analyzing medical images.

\begin{table*}[ht!]
\centering
\caption{Comparison of Method Performance on Graph-Level Attack}
\label{tab:results}
\begin{tabular}{|c|c|c|c|}
\hline
Method                & Proposed Method & GNN (Defferrard et al.) & GAT (Velickovic et al.) \\ \hline
Dataset               & 0.8134          & 0.7688                  & 0.7718                  \\ \hline
10\% Perturbed Graphs & 0.7927          & 0.7607                  & 0.7663                  \\ \hline
25\% Perturbed Graphs & 0.7902          & 0.7519                  & 0.7557                  \\ \hline
50\% Perturbed Graphs & 0.7808          & 0.7343                  & 0.7449                  \\ \hline
\end{tabular}
\end{table*}

\begin{table*}[ht!]
\centering
\caption{Comparison of Method Performance on Image-Level Attack}
\label{tab:results2}
\begin{tabular}{|c|c|c|c|c|}
\hline
Method                & Proposed Method & GNN (Defferrard et al.) & GAT (Velickovic et al.) & ViTs (Ghaffari et al.) \\ \hline
Dataset               & 0.8134          & 0.7688                  & 0.7718                  & 0.7923                 \\ \hline
10\% Perturbed Graphs & 0.7864          & 0.7525                  & 0.7613                  & 0.7816                 \\ \hline
25\% Perturbed Graphs & 0.7858          & 0.7456                  & 0.7544                  & 0.7698                 \\ \hline
50\% Perturbed Graphs & 0.7789          & 0.7381                  & 0.7431                  & 0.7612                 \\ \hline
\end{tabular} 
\end{table*}

\section{Implementation Details}

For generating adversarial samples at the image level, we utilized the FGSM, a widely recognized technique for creating perturbations that can affect model performance. FGSM works by making small, calculated modifications to the original images. To implement this, we used a ResNet50 model pre-trained on the ImageNet dataset as the base architecture. Given that ResNet50 is not designed to process the large size of WSIs, we adapted the proposed approach. We divided each WSI into smaller patches of $256\times256$ pixels, retaining the specific location of each patch. This step was crucial to ensure that we could accurately target each patch with the adversarial attack and then effectively reassemble the WSI from these patches. By attacking these smaller, manageable patches, we were able to apply the FGSM method effectively, despite the large size of the original WSIs. Once each patch was subjected to the adversarial attack, we reconstructed the full WSI from the altered patches. This method allowed us to generate adversarially perturbed versions of the WSIs while overcoming the size limitations of using the ResNet50 model. This approach ensures that we can evaluate the robustness of the model against adversarial threats. 
To generate adversarial samples at the graph level, we employed the meta-gradient technique, which is notably acknowledged in the field of graph-based adversarial learning. Meta-gradient methods involve calculating gradients with respect to the graph structure itself, rather than the weights of the network. This approach allows us to create subtle yet effective perturbations directly in the graph's structure, altering the connections and relationships between nodes in a manner that is intended to mislead the GNN.

Prior to applying the proposed method, the WSIs needed to be preprocessed for compatibility with the graph-based algorithm. This involves segmenting each WSI into non-overlapping patches, each measuring $256\times256$ pixels at $16 \times$ resolution. The features for these patches were extracted using CTranspath~\cite{wang2022}. The number of nodes in the graph was based on the number of patches in each WSI. 

The feature extraction phase was followed by the training of the model using the original dataset for 60 epochs. For this training process, we employed the Adam optimizer, recognized for its efficiency in handling large datasets and complex architectures. The choice of loss function was cross-entropy, a standard in classification tasks due to its effectiveness in penalizing incorrect classifications. To optimize the learning process, we set the learning rate at $1e-3$  and applied a weight decay of $5e-5$. These parameters were meticulously chosen to balance the speed of convergence with the need to avoid overfitting, ensuring that the model learns generalizable patterns from the data. Subsequently, the model was trained on perturbed samples, conducted with and without the implementation of a denoiser. This approach allowed us to compare the model's performance under varying conditions and assess the effectiveness of the denoising process. 
The development and training of the model were conducted using PyTorch and Pytorch\_geometric, two powerful DL frameworks that offer flexibility and extensive support for graph-based data. The computational training was performed on an NVIDIA Tesla V100 GPU. This hardware setup ensured efficient training and model optimization, facilitating a robust analysis of the model performance under different conditions.

For the evaluation phase, we utilized two widely recognized metrics in ML and statistics: the kappa score and accuracy. The kappa score provides insight into the agreement between the model's predictions and the actual labels, adjusting for the possibility of random agreement. Meanwhile, accuracy offers a straightforward measure of the model's overall correctness in classification.

\section{Numerical Results}

To evaluate the effectiveness of the proposed model, we performed a comparative analysis with major graph-based architectures, including GNN and GAT. The focus was on these methods due to the lack of research on enhancing the robustness of graph-based learning for WSIs. Additionally, we benchmarked against the advanced patch-based method, ViTs, known for enhancing WSIs' robustness, especially for image-based attacks. The assessment focused on how these methods fare against adversarial attacks, considering situations in which both attacks occur on the image or graph level. Initially, we assessed the proposed model's performance under standard conditions, free from any attack interference. The findings indicate that the proposed method outperforms both CNN patch-based and graph-based methods in this scenario. The accuracy of the model without any attack is 81. 34\%, and its Kappa score is 0.92. Subsequently, we introduced attacked samples to the dataset at both the image and graph levels, at rates of 10\%, 25\%, and 50\%. For these attacks, the WSI and the graphs were randomly selected. The results for each type of attack, both at the image level and the graph level, are comprehensively detailed in Table \ref{tab:results} and Table \ref{tab:results2}, respectively. In these tables, we provide an in-depth analysis of the results, including a comparison for each specific attack scenario. This detailed presentation allows for a clearer understanding of how the proposed model responds to  types of attacks, offering insights into its performance and robustness in varied conditions. Table \ref{tab:results} illustrates that the model's accuracy decreased by 2.07\%, 2.32\%, and 3.26\% correspondingly when the dataset was subjected to attack rates of 10\%, 25\%, and 50\% in graphs. Furthermore,  Table \ref{tab:results2} illustrates that the model's accuracy decreased by 2.7\%, 2.76\%, and 3.45\% correspondingly when the dataset was subjected to attack rates of 10\%, 25\%, and 50\% in images.

\section{Conclusion}
The significance of adversarial attacks and the enhancement of the robustness of DL models are crucial factors in establishing trust in the deployment of these technologies, particularly in sensitive domains such as healthcare. Given the significant financial incentives in the healthcare domain, the occurrence of adversarial attacks is an inevitable challenge that must be addressed to ensure the safe and reliable use of these technologies. In this research, we introduced a novel approach designed to enhance the robustness of cancer Gleason classification against adversarial attacks, both at the image and graph levels.

In order to evaluate the effectiveness of the proposed model, we conducted a comparative analysis against leading graph-based architectures like GNN and GAT. Given the lack of existing research on enhancing the robustness of graph-based learning specifically for WSIs, the focus was directed towards the most prominent graph-based methods. We aimed to analyze how these methods respond to perturbation in both at image and graph levels, in the presence and absence of the novel denoising approach. the evaluation process began by testing the performance of the model under standard conditions, without the introduction of any perturbations. This initial evaluation provided a baseline understanding of the model's capabilities in processing and analyzing WSIs. It was crucial to establish this benchmark to appreciate the impact of subsequent adversarial challenges and the effectiveness of the denoising strategy.

The approach represents a significant innovation, particularly due to its design that targets adversarial attacks at the WSI level, moving beyond the constraints of traditional patch-based methods. It delves into various methods and locations where adversarial attacks might occur. By concentrating on WSI, this research offers a comprehensive and precise examination of the diverse variabilities found in computer-aided cancer detection, setting a groundbreaking precedent in the field.

\bibliographystyle{IEEEtran}
\bibliography{tmi}
\end{document}